\documentclass{article} 

\PassOptionsToPackage{numbers, compress}{natbib}

\usepackage[preprint]{neurips_2025}

\usepackage{amsmath,amsfonts,bm}

\def\eqref#1{equation~\ref{#1}}

\def\1{\bm{1}}

\DeclareMathAlphabet{\mathsfit}{\encodingdefault}{\sfdefault}{m}{sl}
\SetMathAlphabet{\mathsfit}{bold}{\encodingdefault}{\sfdefault}{bx}{n}

\usepackage[utf8]{inputenc}
\usepackage[T1]{fontenc}
\usepackage{hyperref}
\usepackage{url}
\usepackage{enumerate}
\usepackage{booktabs}
\usepackage{graphicx}
\usepackage{subcaption}
\usepackage{multirow}
\usepackage{amsmath}
\usepackage{amssymb}
\usepackage{tikz}
\usetikzlibrary{arrows.meta, positioning, calc, decorations.pathreplacing}

\title{Beyond the Next Step: Variable-Length Latent World Models for Long-Horizon Planning}

\author{Tianqi Du\textsuperscript{1*} \qquad Qi Zhang\textsuperscript{1}\thanks{Equal Contribution.} \qquad Yifei Wang\textsuperscript{2}\qquad Yisen Wang\textsuperscript{1,3}\thanks{Corresponding Author: Yisen Wang (yisen.wang@pku.edu.cn).}\quad \\ \textsuperscript{1} State Key Lab of General Artificial Intelligence,\\ \hspace{0.14cm} School of Intelligence Science and Technology, Peking University\\  \textsuperscript{2} Amazon AGI SF Lab \\ \textsuperscript{3} Institute for Artificial Intelligence, Peking University}

\begin{document}

\maketitle

\begin{abstract}
Recently, world models have emerged as a promising paradigm for building intelligent agents by learning predictive models that estimate future environment states conditioned on observations and actions. In particular, JEPA-style latent world models provide an efficient alternative to pixel space prediction by learning action-conditioned dynamics in compact representation spaces. However, existing latent world models typically rely on one-step prediction and must be recursively rolled out for long-horizon planning, which leads to compounding errors and a mismatch between training objectives and downstream planning tasks. To address this limitation, we propose Variable-length Latent World Models (VLWMs), a framework that learns to predict future latent states conditioned on action sequences of variable lengths. Instead of training only on one-step transitions, VLWMs directly model temporally extended dynamics, allowing the same predictor to evaluate action plans over different horizons. We further introduce a curriculum training strategy that progressively expands the action horizon, stabilizing optimization from short-range dynamics to long-range prediction. At test time, we design planning methods tailored to VLWMs to better exploit their variable-length predictive capabilities. Experiments on long-horizon control tasks show that VLWMs significantly improve latent space world models, achieving 13\% average improvement over the state-of-the-art LeWM across different datasets, with especially large gains on tasks requiring extended planning. These results suggest that VLWM provides a simple yet effective paradigm for improving long-horizon prediction and planning in latent world models.

\end{abstract}

\section{Introduction}

World models have emerged as a powerful paradigm for visual decision-making and control, where agents learn environment dynamics by predicting future states from past observations and actions, enabling planning through imagined rollouts \citep{ha2018world}. Despite sharing this common objective, world models encompass a diverse range of formulations and implementations, differing in their prediction targets, learning objectives, and representation spaces \citep{ding2025understanding,bardes2023v,bruce2024genie,wan2025wan}. Among them, recent self-supervised approaches, such as JEPA-style predictive architectures \citep{bardes2023v,assran2025v} and DINO-WM \citep{zhou2024dino}, have demonstrated that predicting future states in a learned latent space, rather than reconstructing pixels, yields highly efficient simulators for downstream planning and control. These latent world models have achieved impressive performance across robotic manipulation, navigation, and visual control \citep{maes2026leworldmodel,hansen2024td}. 

Despite their success on short-horizon tasks, existing world models still struggle severely with \textbf{long-horizon} planning, where rollout errors accumulate over time, leading to temporal drift and rapidly deteriorating trajectory quality \citep{villegas2017learning,zhang2026hierarchical}. We argue that this is not merely an engineering issue but a consequence of how these models are trained. Most world models are trained with a fixed-step prediction objective, which provides supervision only at a local temporal scale. Consequently, long-range consistency is never explicitly optimized and the training signal is mismatched with the long-horizon nature of the planning problem. Furthermore, videos and ego-centric observations exhibit substantial temporal redundancy, with adjacent frames often containing highly correlated information \citep{pan2021va}. As a result, next-step prediction can frequently be solved through short-term interpolation or near-identity mappings, encouraging models to capture local appearance changes rather than the higher-level dynamics required for long-horizon reasoning and planning.

\begin{figure}[t]
\centering
\begin{minipage}[c]{0.93\linewidth}
\centering
\includegraphics[width=\linewidth]{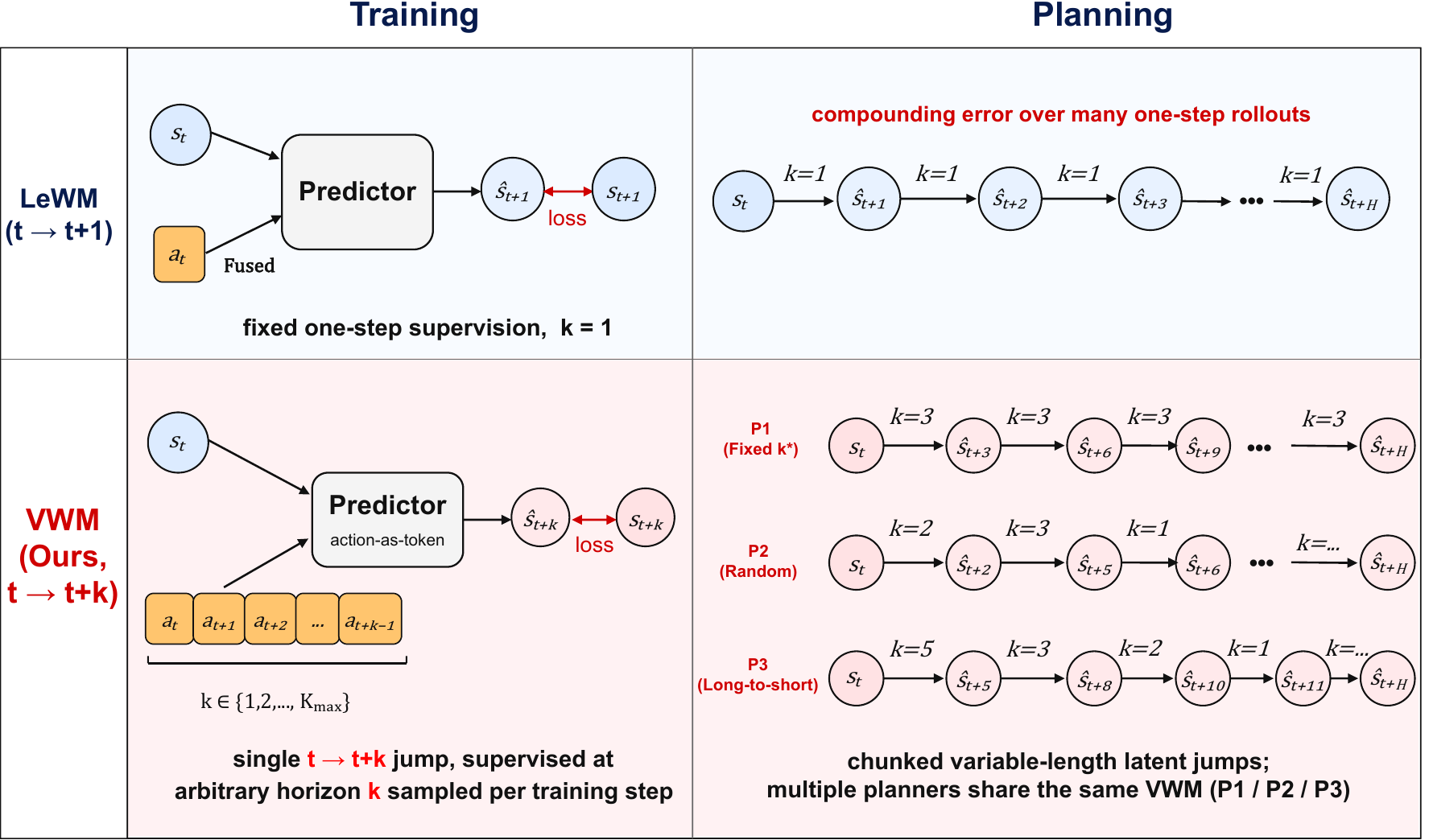}
\end{minipage}
\caption{\textbf{Overview of Variable-length Latent World Models (VLWM).} Compared to a one-step world model (LeWM, top), VLWM (bottom) trains a single predictor to forecast the latent at an arbitrary offset $t{\to}t{+}k$ by feeding a variable-length segment of action tokens, and at planning time replaces the long autoregressive rollout with a few \textbf{chunked} variable-length latent jumps, mitigating compounding error.}
\label{fig:teaser}
\end{figure}

To mitigate this limitation, we propose Variable-length Latent World Models (VLWM), a framework that enables a single model to predict future states at variable horizons. Rather than learning to forecast at a fixed step size, VLWM is trained to predict latent states at $t+k$ for varying values of $k$, which provides a unified learning objective that simultaneously captures short-term dynamics and long-range semantic consistency. Under this perspective, existing latent world models such as JEPA \citep{bardes2023v} and DINO-WM \citep{zhou2024dino} can be interpreted as special cases of VLWM corresponding to fixed prediction horizons. A second key component in VLWM is a unified token-based formulation for actions. Unlike prior world models that incorporate actions through dedicated conditioning mechanisms (e.g., FiLM layers or cross-attention modules), VLWM treats each action embedding as a standard transformer token and interleaves action and state tokens within a single sequence. This formulation eliminates the need for a predefined action-conditioning interface, allowing the model to naturally condition on an arbitrary number of past actions while preserving architectural simplicity and flexibility.

A key challenge in learning variable-length latent world models is that training with large prediction horizons from the outset is highly unstable. Long-horizon prediction tasks are substantially more difficult, often resulting in noisy gradients, optimization failure, and representation collapse. To address this, VLWM adopts a curriculum strategy that gradually expands the prediction horizon, allowing long-range semantics to be built upon well-learned short-horizon dynamics. Together with a balanced horizon sampling scheme and a shared encoder-predictor architecture, this enables stable and scalable variable-length training. Beyond training, VLWM also provides a more flexible planning interface: planners can adapt the prediction horizon to the task, using short horizons for fine-grained control and long horizons to reduce error accumulation over extended trajectories. By combining multiple horizons within a single rollout, VLWM further enables hierarchical planning without requiring separate high-level and low-level policies \citep{zhang2026hierarchical}. Consequently, we believe that variable-length supervision in VLWM provides both an effective learning objective for long-horizon tasks and a foundation for a new class of planning algorithms.

We evaluate our framework on three planning benchmarks spanning short- and long-horizon tasks. Compared to strong world-model baselines such as LeWorldModel \citep{maes2026leworldmodel}, our approach delivers substantial gains on long-horizon planning while matching or improving short-horizon performance, and our analysis shows that variable-length supervision yields measurably better embeddings in terms of long-range predictability and downstream controllability. Our contributions can be summarized as: 
\begin{itemize}
\item We introduce Variable-length Latent World Model (VLWM), a unified framework that learns latent environment dynamics across multiple prediction horizons while naturally supporting variable-length action sequences via an action-as-token formulation.
\item We propose a scalable curriculum learning strategy for stable variable-length training and develop a suite of planning algorithms that adaptively leverage different prediction horizons within a single model.
\item We validate VLWM on a range of planning benchmarks, showing consistent gains over fixed-horizon world models, and provide analyses that reveal the benefits of variable-length supervision for long-horizon planning.
\end{itemize}

\section{Related Work}

\paragraph{World model.}

World models refer to learned models that can predict future states of environments conditioned on observations and actions, enabling planning, decision-making, and policy optimization of agents. Early approaches primarily operate at the pixel level, directly modeling high-dimensional observations such as images or videos \citep{hafner2019dream,bruce2024genie,wan2025wan,team2025hunyuanworld}. However, pixel-level modeling is often computationally expensive due to the high dimensionality of visual observations. To improve efficiency, recent work has explored learning dynamics in latent spaces, among which joint-embedding predictive architectures (JEPAs) have emerged as a prominent paradigm for learning predictive representations without explicit pixel reconstruction \citep{bardes2023v,assran2025v,zhou2024dino}. Building on this idea, a range of recent works further develop JEPA-style world models for control and planning. For example, LeWorldModel \citep{maes2026leworldmodel} simplifies the loss design and proposes an end-to-end training framework, Causal-JEPA \citep{nam2026causal} incorporates object-centric structure into latent predictive representations to improve generalization, and Think-JEPA \citep{zhang2026thinkjepa} enhances the performance by integrating large vision-language models.

\paragraph{Long horizon planning.}

Long-horizon tasks are inherently challenging for world models, as iterative prediction tends to accumulate errors over time, leading to performance degradation during extended rollouts \citep{villegas2017learning,zhang2026hierarchical,team2025longcat}. To mitigate this issue, a line of work in pixel-level world models focuses on reducing the train–test discrepancy that arises in long-horizon generation. For example, \citep{huang2026self} trained the model on its own generated frames rather than ground-truth sequences, thereby aligning the training distribution with the inference-time rollout distribution. This leads to improved stability over long-horizon generation and reduces error accumulation. While in latent world modeling, the concurrent work Hierarchical World Models (HWM) \citep{zhang2026hierarchical} is the most closely related work to ours, as it also explores the different lengths of action sequences. However, there are three key differences between our approach and HWM: (1) HWM relies on a fixed-length of action sequences during the evaluation, whereas our method supports variable-length planning; (2) HWM generates intermediate high-level goals that are not always guaranteed to be executable, while our method directly models valid transitions; (3) HWM requires an additional high-level predictor, whereas our approach is end-to-end and does not rely on auxiliary modules.

\section{Variable-length Latent World Model}

\subsection{Preliminaries}
\label{sec:prelim}

Recently, a growing line of work has shifted from pixel-level video prediction to latent world modeling \citep{bardes2023v,assran2025v,zhou2024dino,maes2026leworldmodel}, where observations are encoded into compact latent representations and future latent states are predicted conditioned on these representations and the agent's actions. By avoiding the need to reconstruct high-dimensional observations, latent world models can focus their capacity on decision-relevant dynamics, leading to more efficient and effective planning. In this framework, planning is performed directly in latent space by optimizing predicted trajectories. We adopt this latent world modeling paradigm as our starting point and briefly review its standard formulation before discussing its limitations.

\paragraph{Training objectives of latent world models.} Most recent latent world models can be viewed as a unified predictive learning framework consisting of an online encoder $f_\theta$, a target encoder $\bar f_{\bar\theta}$ (typically an EMA copy of $f_\theta$), and a predictor $g_\phi$. Consider an agent interacting with an environment, producing observations $\{o_t\}_{t=0}^{T}$ and actions $\{a_t\}_{t=0}^{T-1}$. Let $s_t=f_\theta(o_t)$ denote the online latent representation and $e_t=h_\psi(a_t)$ the action embedding. The objective is to predict the latent representation of the next observation conditioned on the current latent state and action:
\begin{equation}
    \mathcal{L} \;=\; \mathbb{E}_t\Big[\, D\!\big(\, g_\phi(s_{\le t},\, e_t)\,,\; \mathrm{sg}\big(\bar{f}_{\bar\theta}(o_{t+1})\big) \,\big) \,\Big] \;+\; \lambda\,\mathcal{R}(s),
    \label{eq:wm-common}
\end{equation}
where $D$ is a distance in latent space, $\mathrm{sg}(\cdot)$ is stop-gradient, and $\mathcal{R}$ is an optional anti-collapse regularizer. Different recent world models can be viewed as instantiations of Eq.~\ref{eq:wm-common} that differ mainly in (i) what plays the role of $f_\theta / \bar{f}_{\bar\theta}$, (ii) the form of $D$, and (iii) whether $\mathcal{R}$ is needed \citep{bardes2023v,assran2025v,zhou2024dino,maes2026leworldmodel}.

Within this formulation, the action enters the model through the predictor as auxiliary conditioning. In current world models, this is usually implemented by fusing the action embedding $e_t$ inside transformer layer through additive conditioning on the hidden states \citep{maes2026leworldmodel}, i.e., the action is wired into the architecture of each block rather than placed in the input sequence itself. This choice is convenient when there is exactly one action per prediction step, but, as we argue next, it tightly couples the model to a fixed action layout.

\paragraph{Planning with latent world models.}
During the evaluation process, a trained latent world model is used as a learned simulator inside a model predictive control (MPC) loop in latent space. At each replanning step, given the current latent $s_t = f_\theta(o_t)$ and a goal observation $o_g$ encoded as $s_g = f_\theta(o_g)$, the planner searches for an action sequence $\mathbf{a} = (a_t, a_{t+1}, \dots, a_{t+H-1})$ of length $H$ that minimizes a terminal cost in latent space:
\begin{equation}
\mathbf{a}^{\star}
=
\arg\min_{\mathbf{a}}
\left\|
\hat{s}_{t+H}(\mathbf{a}) - s_g
\right\|_2^2.
\label{eq:plan-cost}
\end{equation}
where $\hat{s}_{t+H}(\mathbf{a})$ denotes the latent state at time $t+H$ obtained by recursively rolling out the world model under the candidate action sequence $\mathbf{a}$. The optimization in Eq.~\ref{eq:plan-cost} can be solved using either zeroth- or first-order methods, throughout this work, we adopt the Cross-Entropy Method (CEM). This latent-space MPC framework is applied in the evaluation process of most recent latent world models, including LeWM \citep{maes2026leworldmodel}, V-JEPA-2 \citep{assran2025v}, and our proposed VLWM. Within this shared planning framework, the primary distinction lies in how the future latent state $\hat{s}_{t+H}(\mathbf{a})$ is generated.

\subsection{From One-Step Prediction to Variable-Length Horizon Modeling}
\label{sec:VLWM-formulation}

The standard formulation above exhibits two coupled limitations: \textbf{(i) Single-step supervision.} Eq.~\ref{eq:wm-common} supervises only the one-step transition $t \to t{+}1$, resulting in a low-information target for redundant video data and providing no explicit long-horizon training signal, which limits long-horizon planning \citep{villegas2017learning,zhang2026hierarchical}. \textbf{(ii) Architecture-locked action conditioning.} Because actions are fused inside transformer layers, if we provide multiple actions, they will be compressed into shared hidden representations, which makes it difficult to disentangle their individual contributions to the predicted dynamics. Consequently, the architecture itself dictates how many actions can be supplied. In the following, we introduce the Variable-length Latent World Model (VLWM), which directly addresses these two limitations.

\paragraph{Variable-length supervision for latent prediction.}
Instead of always supervising the next step, we train a single predictor $g_\phi$ to forecast the latent at an arbitrary future offset $k\ge 1$. Given the current latent and the action segment driving the agent forward, our objective is
\begin{equation}
    \mathcal{L}_{\text{VLWM}} = \mathbb{E}_{t,\, k \sim p(k)}\Big\| g_\phi\big(s_t,\, a_{t:t+k-1}\big) - \mathrm{sg}\big(\bar{f}_{\bar\theta}(o_{t+k})\big) \Big\|_2^2,
    \label{eq:VLWM}
\end{equation}
where $s_t = f_\theta(o_t)$, $p(k)$ is a distribution over different horizons (specified in Sec.~\ref{sec:VLWM-training}) and $a_{t:t+k-1}$ is the variable-length action segment of length $k$ that drives the latent from time $t$ to time $t{+}k$. We note that the current world model is a special case of Eq.~\ref{eq:VLWM} with $p(k) = \delta_{k=1}$. In the next part, we will present the architectural modifications to latent world models that enable this objective.

\paragraph{Variable-length action conditioning via tokenization.}
To natively support variable-length action sequences, we depart from the layer-internal conditioning in current latent world models. We embed each action with a shared head $e_\tau = h_\psi(a_\tau)$ and treat $e_\tau$ as a \textbf{first-class transformer token}, placed alongside state tokens in a single input sequence:
\begin{equation}
    \mathbf{X} \;=\; \big[\, s_{t-L+1},\; \dots,\; s_{t-1},\; s_t,\; e_t,\; e_{t+1},\; \dots,\; e_{t+k-1} \,\big].
\end{equation}
The sequence consists of an $L$-frame history of state tokens (with $L\ge 1$) ending at the current state $s_t$, followed by the $k$ action tokens that describe how the agent acts between $t$ and $t{+}k$. The transformer is causal, and we read out the predicted latent $\hat{s}_{t+k}$ directly from the output of the last token in the sequence. No auxiliary $[\textsc{Pred}]$ or mask token is needed. State and action tokens are distinguished by additive type embeddings, and a single learnable absolute position embedding is shared across all horizons. Because the sequence length is no longer hard-wired, the same predictor accepts an arbitrary length of future action segments and outputs the corresponding $\hat{s}_{t+k}$ without any architectural change.

Consequently, VLWM provides three desirable properties at once: (1) supervision at any horizon $k$, (2) variable-length action conditioning by construction, and (3) backbone agnosticism---the encoder $f_\theta$ can be a JEPA-style, DINO-style, or any other latent encoder, and prior fixed-horizon world models are recovered as special cases of $p(k)$.

\subsection{A Curriculum Training Strategy of VLWM}
\label{sec:VLWM-training}

Directly optimizing Eq.~\ref{eq:VLWM} with a broad length of action sequences from scratch is unstable in practice. Long-horizon prediction targets are inherently more uncertain, leading to noisy gradient signals for the predictor. We therefore adopt a \textbf{curriculum} strategy, which gradually introduces longer-range targets once the model has sufficiently learned shorter-horizon predictions.

Specifically, let $T$ denote the total number of training steps and $K_{\max}$ the maximum prediction horizon. We divide training into $K_{\max}$ stages of equal length $T / K_{\max}$, and assign each stage with an increasing index:
\begin{equation}
    j(\tau) \;=\; \min\!\Big(K_{\max},\; \big\lceil K_{\max}\,\tau / T \big\rceil\Big), \qquad \tau \in \{1,\dots,T\}.
\end{equation}
Within stage $j(\tau)$, we sample the horizon $k$ \textbf{uniformly} from the integers $\{1, 2, \dots, j(\tau)\}$, i.e.,
\begin{equation}
    p_\tau(k) \;=\; \mathrm{Uniform}\!\big(\{1,2,\dots,j(\tau)\}\big) \;=\; \frac{1}{j(\tau)}\,\mathbb{1}\!\big[\,1 \le k \le j(\tau)\,\big].
    \label{eq:curriculum}
\end{equation}
Concretely, in the first stage, the model is trained only on $k=1$. In the second stage, we introduce $k=2$ and sample uniformly from ${1,2}$. In general, the $j$-th stage samples $k$ uniformly from ${1,\dots,j}$, and the final stage covers the full range ${1,\dots,K_{\max}}$. Under this cumulative-uniform schedule, newly introduced horizons receive immediate supervision, while previously learned horizons are continually revisited, helping preserve short-horizon dynamics. We further compare against direct training with the full horizon range ${1,\dots,K_{\max}}$ (i.e., without curriculum) in Sec.~\ref{sec:experiments}, and find that the curriculum is essential for stable variable-length training.

\subsection{Planning with Variable-length Latent World Models}
\label{sec:VLWM-planning}

Within the receding-horizon MPC framework introduced in Sec.~\ref{sec:prelim} (Eq.~\ref{eq:plan-cost}), the crucial difference between our planner and a standard MPC controller lies in how $\hat{s}_{t+H}(\mathbf{a})$ is computed. Rather than committing to a fixed step size, we partition the planning horizon $H$ into contiguous chunks $H = k_1 + k_2 + \dots + k_M$ and evaluate the rollout as a sequence of \textbf{variable-length latent jumps}:
\begin{equation}
    s_t \xrightarrow{k_1} \hat{s}_{t+k_1} \xrightarrow{k_2} \hat{s}_{t+k_1+k_2} \xrightarrow{k_3} \cdots \xrightarrow{k_M} \hat{s}_{t+H},
\end{equation}
i.e., the predictor takes the current latent and the full $k_i$-step action segment and directly predicts the latent $k_i$ steps ahead in one shot (cf.\ Eq.~\ref{eq:VLWM}). Standard one-step rollout is recovered as the special case $k_1 = \cdots = k_M = 1$. Larger $k_i$ leverage the multi-horizon supervision introduced in Sec.~\ref{sec:VLWM-formulation} and reduce the number of recursive applications of $g_\phi$. Different chunking schedules ${k_i}$ thus induce different planners built on top of the \textbf{same} VLWM:

\paragraph{(P1) Fixed but task-adaptive horizon.}
The simplest variant uses a constant chunk size $k_i \equiv k^\star$ throughout the rollout and selects $k^\star$ per task: a small $k^\star$ (or even $k^\star{=}1$) for short-horizon, fine-grained control where every action must be evaluated precisely, and a larger $k^\star$ for long-horizon goals where coarse jumps reduce error accumulation.

\paragraph{(P2) Random-horizon sampling.}
At each chunk boundary, we sample $k_i$ uniformly from $\{1,\dots,K_{\max}\}$, with the constraint that $\sum_i k_i = H$ enforced by clipping. This is a stochastic search over the rollout granularity that requires no schedule design and serves as a strong, hyperparameter-light baseline within our framework.

\paragraph{(P3) Long-to-short.}
We start the rollout with large chunks to quickly cover long-range structure between $s_t$ and the goal, and progressively shrink $k_i$ toward the end of the horizon to refine fine-grained dynamics near the predicted terminal state---an analogue of coarse-to-fine search realized inside a single world model. Concretely, we draw $k_i \in \{1,\dots,K_{\max}\}$ from a position-dependent softmax that places more mass on large $k$ early in the horizon and on small $k$ near the end, controlled by a single bias-strength hyperparameter.

All three planners are drop-in: they share the same VLWM weights, the same MPC objective in Eq.~\ref{eq:plan-cost}, and the same solver, and differ only in how the chunk schedule $\{k_i\}$ is chosen. This highlights that variable-length training is not just a better objective but also a planning interface that subsumes a range of fixed- and hierarchical-horizon strategies \citep{zhang2026hierarchical} within a single trained model.

\section{Experiments on Planning Tasks}
\label{sec:experiments}

\subsection{Datasets and Environments}
\label{sec:datasets}

\paragraph{Environments.}
We evaluate VLWM on three goal-conditioned planning environments that span 2D navigation, 2D manipulation, and 3D robotic manipulation, with continuous action spaces and image-based observations. We use the same data and evaluation protocol as \citep{maes2026leworldmodel,zhou2024dino} so that our results are directly comparable to prior latent world models. Further details on data generation and evaluation budgets are deferred to App.~\ref{appendix:envs}.

\begin{enumerate}[a)]
\item \textbf{PushT} is a continuous 2D manipulation task in which an agent (represented as a blue dot) must push a T-shaped block toward a target configuration, with interactions restricted to pushing actions. We follow the same setup and dataset as \citep{zhou2024dino}, which contains $20{,}000$ expert episodes with an average length of $196$ steps.

\item \textbf{OGBench-Cube} is a continuous 3D robotic manipulation task in which a robotic arm with an end-effector must pick up a cube and place it at a target location. We use the single-cube variant introduced by \citep{park2025ogbench} and follow \citep{maes2026leworldmodel} in collecting $10{,}000$ episodes of $200$ steps each, generated by the data-collection heuristic provided in the benchmark library.

\item \textbf{TwoRoom} is a continuous 2D navigation task. The environment consists of two rooms separated by a wall with a single door connecting them. The agent (a red dot) must navigate from a random starting position in one room to a randomly sampled target location in the other room, which requires passing through the door. We collect $10{,}000$ episodes with an average trajectory length of $92$ steps using a noisy heuristic policy that first directs the agent toward the door and then toward the target once the agent has crossed into the other room.
\end{enumerate}

\begin{figure}[t]
\centering
\begin{subfigure}[t]{0.32\linewidth}
    \includegraphics[width=\linewidth]{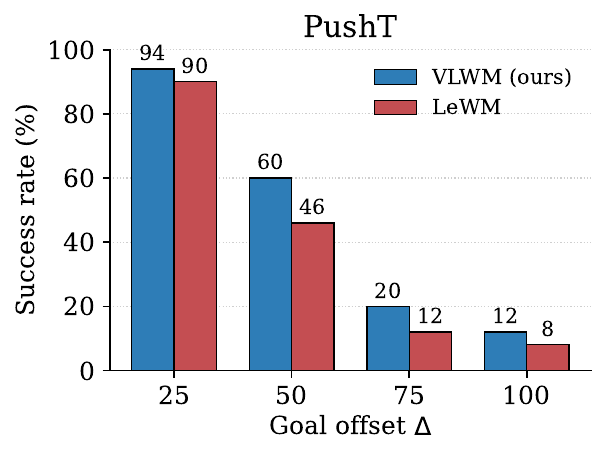}
\end{subfigure}\hfill
\begin{subfigure}[t]{0.32\linewidth}
    \includegraphics[width=\linewidth]{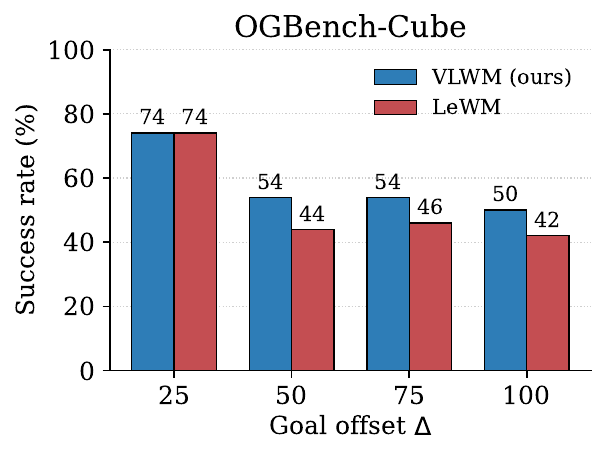}
\end{subfigure}\hfill
\begin{subfigure}[t]{0.32\linewidth}
    \includegraphics[width=\linewidth]{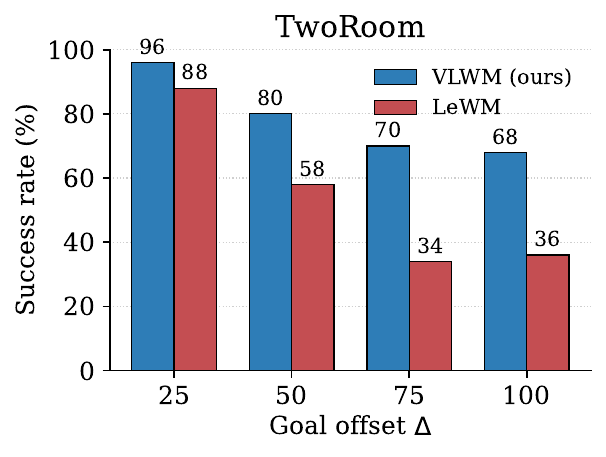}
\end{subfigure}
\caption{Goal-conditioned planning success rate ($\%$) on PushT, OGBench-Cube, and TwoRoom with different offsets $\Delta\in\{25,50,75,100\}$. For each offset, the left bar is VLWM (ours, oracle planner over $\{\text{P1},\text{P2},\text{P3}\}$) and the right bar is LeWM. VLWM consistently outperforms LeWM, with the performance gap widening as the planning horizon increases.}
\label{fig:main}
\end{figure}

\begin{figure}[t]
\centering
\includegraphics[width=0.92\linewidth]{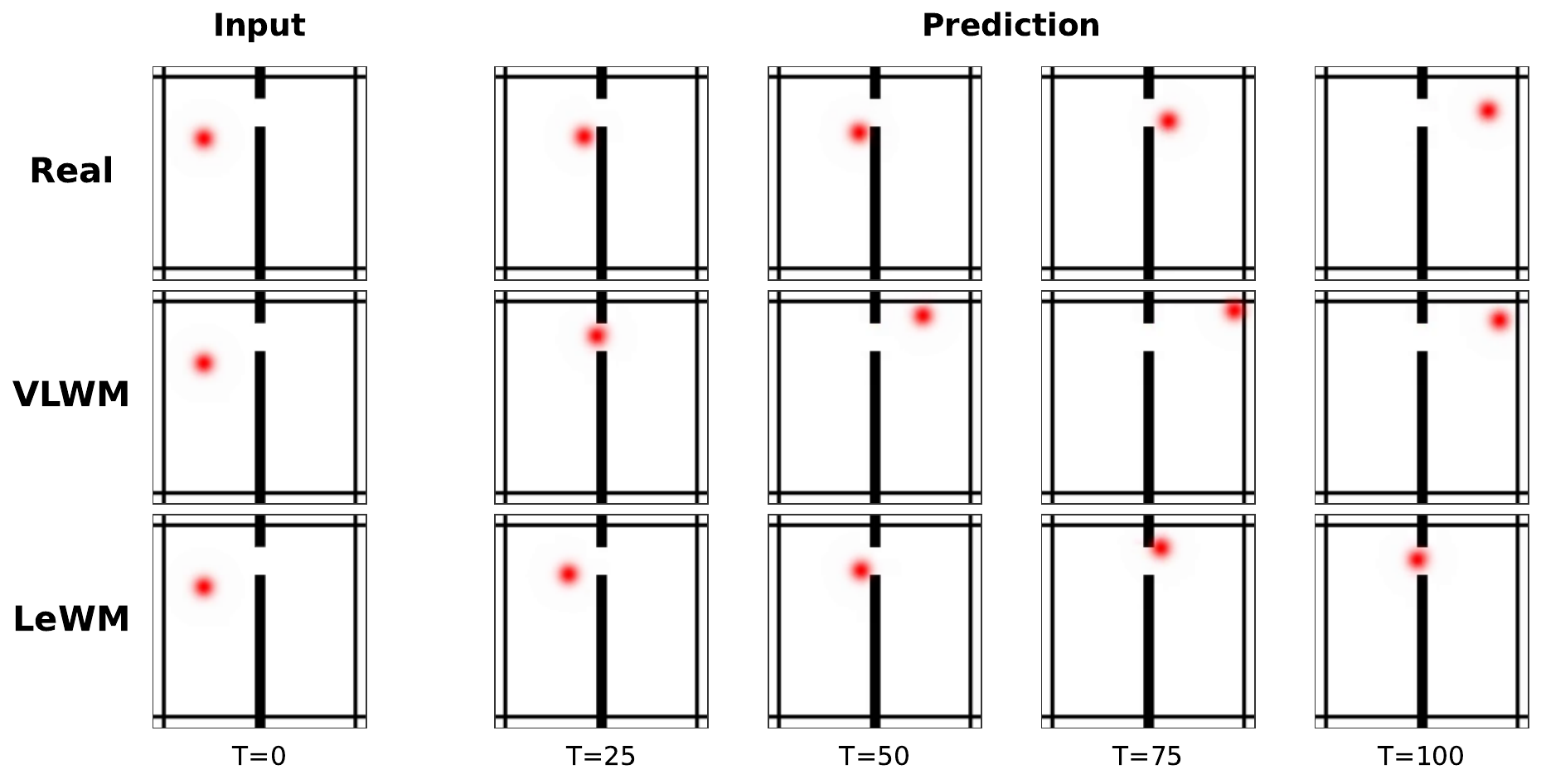}
\caption{\textbf{Visualization of rollouts on TwoRoom.} Given the same input frame at $T{=}0$, we compare VLWM's chunked variable-length rollout (middle) against LeWM's recursive one-step rollout (bottom) at $T\in\{25,50,75,100\}$, with the ground-truth trajectory shown on top. VLWM stays close to the real trajectory and successfully crosses the door into the second room, while LeWM accumulates error and remains stuck in the starting room.}
\label{fig:tworoom-rollout}
\end{figure}

\subsection{Evaluation Protocol}
\label{sec:eval-protocol}

\paragraph{Goal-conditioned planning with a CEM solver.}
We adopt a goal-conditioned planning protocol identical to \citep{maes2026leworldmodel}. For each test episode, we sample an initial observation $o_0$ from a held-out trajectory and define the goal observation $o_g$ to be a frame at \textbf{offset} $\Delta$ steps later in the same trajectory; the agent's task is to drive the environment from $o_0$ to a state whose latent matches $o_g$. Crucially, we vary $\Delta \in \{25, 50, 75, 100\}$ to probe planning quality across short-, medium-, and long-horizon goals. Planning is performed in latent space using the receding-horizon MPC objective in Eq.~\ref{eq:plan-cost}, with a Cross-Entropy Method (CEM) solver applied identically to all methods (same population size, number of iterations, elite ratio, and replanning frequency). Success is determined by an environment-specific criterion (e.g., block configuration error for PushT, end-effector / cube distance for OGBench-Cube, agent-to-target distance for TwoRoom; see App.~\ref{appendix:control_details}), and we report \textbf{success rate} averaged over all evaluation episodes.

\paragraph{Implementation details.}
Our implementation follows \citep{maes2026leworldmodel} closely: encoder backbone, action encoder, predictor depth/width, frame skip, image resolution, batch size, optimizer, and CEM hyperparameters are all kept consistent with the LeWM reference setup unless explicitly noted. The differences specific to VLWM are: (i) the variable-length training objective (Sec.~\ref{sec:VLWM-formulation}); (ii) the action-as-token sequence layout; (iii) the horizon curriculum (Sec.~\ref{sec:VLWM-training}); and (iv) the use of a chunked, variable-length rollout in the CEM cost evaluation (Sec.~\ref{sec:VLWM-planning}). Full hyperparameters and a side-by-side comparison with LeWM's implementation are deferred to App.~\ref{appendix:impl}.

\subsection{Planning Tasks with Different Horizons}
\label{sec:main-results}

Figure~\ref{fig:main} reports goal-conditioned planning success rate on PushT, OGBench-Cube, and TwoRoom for goal offsets $\Delta\in\{25, 50, 75, 100\}$. To avoid cluttering the main comparison with intra-method variants, we present VLWM with an \textbf{oracle planner}---for each (dataset, $\Delta$) configuration we report the best of $\{\text{P1}, \text{P2}, \text{P3}\}$; the full per-planner grid is given in Table~\ref{tab:planner-grid}.

Across all three datasets and all four goal offsets, VLWM with the oracle planner outperforms LeWM. The improvement is largest in the long-horizon regime ($\Delta\in\{75,100\}$), where one-step world models suffer from compounding error: e.g.\ on TwoRoom-$100$ the success rate climbs from $36\%$ (LeWM) to $68\%$ (VLWM); on TwoRoom-$75$ from $34\%$ to $70\%$; on OGBench-Cube-$100$ from $42\%$ to $50\%$. This directly validates our motivation in Sec.~\ref{sec:prelim} that single-step supervision is the bottleneck for long-horizon planning. Figure~\ref{fig:tworoom-rollout} qualitatively illustrates this gap on TwoRoom: starting from the same initial frame, VLWM's predicted latents (decoded for visualisation) closely track the ground-truth trajectory through the door and into the second room across $T\in\{25,50,75,100\}$, whereas LeWM's recursive one-step rollout drifts and never crosses the door, exactly the failure mode the planning numbers reflect.
\begin{table}[t]
\centering
\caption{Success rate ($\%$) of different VLWM planning strategies. P1 uses a fixed chunk size, P2 randomly samples each chunk length from a uniform distribution, while P3 progressively shifts from larger to smaller chunk sizes during rollout using a position-dependent distribution. Bold marks the column-wise best result among VLWM planners.}
\label{tab:planner-grid}
\small
\setlength{\tabcolsep}{4pt}
\begin{tabular}{l cccc cccc cccc}
\toprule
& \multicolumn{4}{c}{PushT} & \multicolumn{4}{c}{OGBench-Cube} & \multicolumn{4}{c}{TwoRoom} \\
\cmidrule(lr){2-5}\cmidrule(lr){6-9}\cmidrule(lr){10-13}
Planner & 25 & 50 & 75 & 100 & 25 & 50 & 75 & 100 & 25 & 50 & 75 & 100 \\
\midrule
LeWM (base, $k{=}1$)              & 90 & 46 & 12 & ~8 & \textbf{74} & 44 & 46 & 42 & 88 & 58 & 34 & 36 \\
P1 fixed $k^\star{=}3$            & 84 & 52 & ~6 & ~2 & 70 & 48 & 52 & 46 & 94 & \textbf{80} & 62 & 58 \\
P1 fixed $k^\star{=}5$            & 88 & 56 & 14 & ~8 & 72 & 44 & 50 & 44 & 90 & 70 & 60 & 62 \\
P2 random ($K_{\max}{=}5$)        & 86 & 52 & 10 & ~6 & \textbf{74} & \textbf{54} & \textbf{54} & \textbf{50} & 92 & 72 & 66 & 52 \\
P3 long-to-short ($K_{\max}{=}3$) & \textbf{94} & \textbf{60} & \textbf{20} & \textbf{12} & 68 & 52 & 48 & 44 & \textbf{96} & 70 & 64 & 60 \\
P3 long-to-short ($K_{\max}{=}5$) & 84 & 48 & ~8 & ~4 & 66 & 46 & 50 & 40 & 90 & \textbf{80} & \textbf{70} & \textbf{68} \\
\bottomrule
\end{tabular}
\end{table}

\begin{table}[t]
\centering
\caption{Success rate ($\%$) of VLWM (oracle planner per cell) with and without the cumulative-uniform curriculum of Eq.~\ref{eq:curriculum}. The curriculum strategy is broadly beneficial: it consistently helps on PushT and TwoRoom. Bold marks the better variant per cell.}
\label{tab:curriculum}
\small
\setlength{\tabcolsep}{4pt}
\begin{tabular}{l cccc cccc cccc}
\toprule
& \multicolumn{4}{c}{PushT} & \multicolumn{4}{c}{OGBench-Cube} & \multicolumn{4}{c}{TwoRoom} \\
\cmidrule(lr){2-5}\cmidrule(lr){6-9}\cmidrule(lr){10-13}
Variant & 25 & 50 & 75 & 100 & 25 & 50 & 75 & 100 & 25 & 50 & 75 & 100 \\
\midrule
VLWM, no curriculum   & 90 & 58 & 14 & ~8 & 72 & \textbf{60} & \textbf{62} & 48 & \textbf{96} & 76 & 68 & 64 \\
VLWM, with curriculum & \textbf{94} & \textbf{60} & \textbf{20} & \textbf{12} & \textbf{74} & 54 & 54 & \textbf{50} & \textbf{96} & \textbf{80} & \textbf{70} & \textbf{68} \\
\bottomrule
\end{tabular}
\end{table}

\subsection{Empirical Understandings}
\label{sec:analysis}

\paragraph{Planner-selection analysis.}
Table~\ref{tab:planner-grid} explores the performance of different planning strategies of VLWMs, revealing two key observations. First, planners that leverage variable-length transitions consistently outperform both LeWM and fixed-step planning across most settings, highlighting the importance of adapting rollout granularity during long-horizon planning. Second, among variable-length planners, no single strategy is universally optimal. While the random planner P2 achieves the best performance in several settings, particularly on OGBench-Cube, the long-to-short planner P3 performs best in others, including all offsets on PushT and the longest offsets on TwoRoom. These results suggest that different environments and planning horizons benefit from different ways of allocating coarse and fine-grained transitions during rollout. Consequently, developing a principled adaptive strategy that selects ${k_i}$ online remains an important direction for future work.

\paragraph{Effect of curriculum training.}
The numbers reported in Figure~\ref{fig:main} use the cumulative-uniform curriculum of Eq.~\ref{eq:curriculum}. Removing the curriculum, i.e.\ training directly on the full range $\{1,\dots,K_{\max}\}$ from the start, generally hurts performance, and the curriculum is at least as good as no-curriculum on every (PushT, TwoRoom) cell (Table~\ref{tab:curriculum}). The gains are most pronounced at long horizons, e.g.\ $+6$ on PushT-$75$ ($14\%\!\to\!20\%$) and $+4$ on PushT-$100$, TwoRoom-$50$, and TwoRoom-$100$, supporting our claim that progressively expanding the horizon stabilises long-range optimisation. The notable exception is OGBench-Cube, where the curriculum offers little advantage and even slightly hurts at mid horizons (e.g.\ $-6$ at $\Delta{=}50$ and $-8$ at $\Delta{=}75$). We hypothesise that this dataset simply does not benefit much from long-horizon supervision in the first place: the LeWM baseline shows essentially flat success rates across $\Delta\in\{50,75,100\}$ ($44/46/42$, Table~\ref{tab:planner-grid}), suggesting that further-out goals on Cube are not appreciably harder than mid-range ones, so the extra long-horizon training signal that the curriculum is designed to scaffold yields limited additional benefit. Crucially, on the two environments where long-horizon planning genuinely matters, the curriculum delivers consistent improvements without changing the qualitative ordering of planners or the gap to LeWM.

\paragraph{Probing physical structure of the latent space.}
A complementary way to assess representation quality, beyond planning success, is to ask which physical quantities can be linearly read off from the learned latents. Following the probing protocol of \citep{maes2026leworldmodel}, we freeze each world model after training, extract the encoder's latent $\bar{f}_{\bar\theta}(o)$ on held-out PushT frames, and fit a linear probe (single fully-connected layer) trained with MSE loss to predict three ground-truth physical quantities: agent location, block location, and block angle. We report MSE ($\downarrow$) and Pearson correlation $r$ ($\uparrow$) on the test split. The probe is trained with identical optimisation hyperparameters, training data, and batch size for VLWM and LeWM, so any difference is attributable to the encoder--predictor itself. Table~\ref{tab:probe-pusht} reports the comparison: VLWM achieves higher Pearson correlation $r$ than LeWM on all three properties, and lower MSE on agent location ($0.076$ vs.\ $0.088$) and block angle ($0.142$ vs.\ $0.169$); on block location the two models are essentially tied, with VLWM slightly higher in MSE but still better in $r$. The largest gains appear on block angle, a property that requires modeling slow semantic state changes rather than instantaneous appearance. We attribute this improvement to VLWM's longer-range supervision, which encourages the encoder to capture richer semantic structure in the scene.

\begin{table}[t]
\centering
\caption{Linear-probe evaluation of latent representations. A linear probe is trained on frozen encoder latents to predict ground-truth physical properties, and evaluated using MSE ($\downarrow$) and Pearson correlation $r$ ($\uparrow$) on a held-out split. VLWM achieves comparable or better performance than LeWM across all properties, with the largest gain on block angle.}
\label{tab:probe-pusht}
\small
\setlength{\tabcolsep}{6pt}
\begin{tabular}{l cc cc cc}
\toprule
& \multicolumn{2}{c}{\textbf{Agent Location}} & \multicolumn{2}{c}{\textbf{Block Location}} & \multicolumn{2}{c}{\textbf{Block Angle}} \\
\cmidrule(lr){2-3}\cmidrule(lr){4-5}\cmidrule(lr){6-7}
\textbf{Model} & MSE $\downarrow$ & $r$ $\uparrow$ & MSE $\downarrow$ & $r$ $\uparrow$ & MSE $\downarrow$ & $r$ $\uparrow$ \\
\midrule
LeWM        & 0.088  & 0.955  & \textbf{0.029}  & 0.975  & 0.169 & 0.882  \\
VLWM (ours)  & \textbf{0.076}  & \textbf{0.971}  & 0.035  & \textbf{0.981}  & \textbf{0.142}  & \textbf{0.904}  \\
\bottomrule
\end{tabular}
\end{table}

\section{Conclusion}

We introduce Variable-length Latent World Models (VLWM), a simple yet effective framework for long-horizon planning in latent world models. In contrast to prior approaches based on fixed one-step prediction and recursive rollout, VLWM learns to predict future latent states at variable temporal offsets conditioned on corresponding action segments, unifying short- and long-horizon dynamics learning within a single objective. To stabilize optimization, we introduce a horizon curriculum that progressively increases prediction difficulty, along with a token-based action representation that removes architectural constraints on action conditioning. On top of this model, we design planning strategies that adaptively adjust chunk sizes during rollout, enabling both fine-grained control and long-range reasoning within a single framework. Experiments demonstrate that VLWM consistently improves long-horizon planning performance, where conventional world models suffer from compounding errors, and further analysis shows that variable-length supervision yields more predictive latent representations and a more flexible planning interface. Overall, we believe that VLWM provides a new direction toward enhancing long-horizon planning ability in latent world models.

\bibliography{iclr2026_conference}
\bibliographystyle{iclr2026_conference}

\newpage
\appendix

\section{Environment and Dataset Details}
\label{appendix:envs}

We adopt the same three environments and datasets as LeWM~\citep{maes2026leworldmodel}, so that the planning protocol, data, and evaluation budgets are directly comparable.

\paragraph{TwoRoom.}
TwoRoom is a simple continuous 2D navigation task introduced by \citep{sobal2025stresstesting}. The environment consists of two rooms separated by a wall with a single door connecting them. The agent (a red dot) must navigate from a random starting position in one room to a randomly sampled target location in the other room, which requires passing through the door. We collect $10{,}000$ episodes with an average trajectory length of $92$ steps. The data are generated using a simple noisy heuristic policy that first directs the agent toward the door along a straight-line path and then toward the target location once the agent has crossed into the other room. Each world model is trained on this dataset for $10$ epochs, matching the LeWM reference setup.

\paragraph{PushT.}
PushT is a continuous 2D manipulation task in which an agent (a blue dot) must push a T-shaped block to match a target configuration, with interactions restricted to pushing actions. We follow the same setup and dataset as \citep{zhou2024dino}, which contains $20{,}000$ expert episodes with an average length of $196$ steps. Each world model is trained on this dataset for $10$ epochs, matching the LeWM reference setup.

\paragraph{OGBench-Cube.}
OGBench-Cube is a continuous 3D robotic manipulation task in which a robotic arm with an end-effector must pick up a cube and place it at a target location. Originally introduced by \citep{park2025ogbench}, we consider only the single-cube variant. We collect $10{,}000$ episodes, each consisting of $200$ steps, using the data-collection heuristic provided in the benchmark library. Each world model is trained on this dataset for $10$ epochs, matching the LeWM reference setup.

\section{Control and Success Criteria}
\label{appendix:control_details}

We follow the goal-conditioned control protocol of LeWM~\citep{maes2026leworldmodel}. Control performance is measured by two parameters: the \emph{evaluation budget}, i.e., the maximum number of actions the agent is allowed to execute in the environment, and the \emph{goal offset} $\Delta$, i.e., how many timesteps in the future the goal state is sampled relative to the initial state. During evaluation, trajectories are sampled from the offline dataset; the initial state is randomly sampled from a held-out trajectory and the goal state corresponds to a state occurring $\Delta$ timesteps later in the same trajectory, which ensures that the goal is reachable and consistent with the dataset dynamics.

In all three environments (TwoRoom, PushT, OGBench-Cube), we set the evaluation budget to $50$ environment steps and vary the goal offset $\Delta \in \{25, 50, 75, 100\}$ to probe planning quality across short-, medium-, and long-horizon goals. Whether a trajectory is counted as successful is determined by an environment-specific criterion (block configuration error for PushT, end-effector / cube distance for OGBench-Cube, agent-to-target distance for TwoRoom), exactly as in LeWM~\citep{maes2026leworldmodel}.

\section{Implementation Details}
\label{appendix:impl}

We follow the implementation of LeWM~\citep{maes2026leworldmodel} for all components that are not specific to variable-length training, and only modify the items required by VLWM. We list the configuration we actually use below; all unspecified hyperparameters match the LeWM reference setup.

\paragraph{Common training setup.}
Each frame is rendered at $224 \times 224$ resolution. We apply a frame-skip of $5$, grouping consecutive actions between frames into a single action block. We train with AdamW (learning rate $5{\times}10^{-5}$, weight decay $1{\times}10^{-3}$), batch size $128$, gradient clipping at norm $1.0$, and bfloat16 mixed precision. Although the optimiser is configured to run for up to $100$ epochs, we always evaluate the checkpoint at the end of epoch $10$, which we empirically find to be sufficient to reach the best planning performance (consistent with the observation in the LeWM reference setup). We use seed $3072$ and a $90/10$ train/validation split of each dataset.

\paragraph{Encoder.}
The encoder $f_\theta$ is a Vision Transformer Tiny (ViT-Tiny) operating on $224 \times 224$ inputs with patch size $14$, projected to an embedding dimension of $192$. The encoder is trained jointly with the predictor (no frozen backbone) and uses ImageNet input normalisation.

\paragraph{Predictor.}
The predictor $g_\phi$ is a $6$-layer causal transformer with $16$ heads, head dimension $64$, MLP hidden dimension $2048$, and dropout $0.1$. Its history window covers the last $3$ state tokens (i.e., $L{=}3$ in Sec.~\ref{sec:VLWM-formulation}). Action embeddings $e_\tau = h_\psi(a_\tau)$ are produced by a shared linear head and are inserted as first-class tokens in the input sequence.

\paragraph{VLWM-specific configuration.}
VLWM departs from LeWM only in: (i) the variable-length training objective of Eq.~\ref{eq:VLWM}, with maximum horizon $K_{\max}{=}5$ (after frame-skip of $5$, this corresponds to a maximum of $25$ environment steps per chunk); (ii) the action-as-token sequence layout described in Sec.~\ref{sec:VLWM-formulation}, which distinguishes state and action tokens through additive type embeddings while sharing a single learnable absolute position embedding across all $k$; (iii) the cumulative-uniform horizon curriculum of Eq.~\ref{eq:curriculum}, split into $K_{\max}$ equal-length stages over the first $10$ epochs of training; and (iv) the chunked, variable-length rollout used in CEM cost evaluation (Sec.~\ref{sec:VLWM-planning}). All other architectural and optimisation choices are identical to LeWM, so any difference in success rate is attributable to the model itself rather than to the training budget or the planner.

\paragraph{Planning solver.}
For planning, we use the Cross-Entropy Method (CEM)~\citep{rubinstein2004cross}. At each replanning step, CEM samples $300$ candidate action sequences from a Gaussian distribution (initial variance $1$) and optimises them for $30$ iterations; at each iteration, the top $30$ trajectories are retained as elites to update the sampling distribution. The planning horizon is $5$ planning steps (corresponding to $25$ environment timesteps under the frame-skip of $5$), and we employ a receding-horizon Model Predictive Control (MPC) scheme with receding horizon $5$, meaning the entire optimised action sequence is executed before replanning. This setup follows \citep{zhou2024dino,maes2026leworldmodel}. The choice of \emph{planner} (P1, P2, P3) further determines how the $5$-step planning horizon is partitioned into chunk sizes $\{k_i\}$ for variable-length rollout, as described in Sec.~\ref{sec:VLWM-planning}.

\end{document}